# Efficient IRIS Recognition Through Improvement of Feature Extraction and subset Selection

*Amir Azizi*
Islamic Azad University Mashhad Branch
Mashhad , IRAN
Amirazizi_b@Yahoo.com

*Hamid Reza Pourreza*
Ferdowsi University of Mashhad
Mashhad , IRAN
hpourreza@um.ac.ir

*Abstract*—the selection of the optimal feature subset and the classification has become an important issue in the field of iris recognition. In this paper we propose several methods for iris feature subset selection and vector creation. The deterministic feature sequence is extracted from the iris image by using the contourlet transform technique. Contourlet transform captures the intrinsic geometrical structures of iris image. It decomposes the iris image into a set of directional sub-bands with texture details captured in different orientations at various scales so for reducing the feature vector dimensions we use the method for extract only significant bit and information from normalized iris images. In this method we ignore fragile bits. And finally we use SVM (Support Vector Machine) classifier for approximating the amount of people identification in our proposed system. Experimental result show that most proposed method reduces processing time and increase the classification accuracy and also the iris feature vector length is much smaller versus the other methods.

*Keywords-Biometric-Iris Recognition-Contourlet-Support Vector Machine (SVM)*

## I. INTRODUCTION

THERE has been a rapid increase in the need of accurate and reliable personal identification infrastructure in recent years, and biometrics has become an important technology for the security. Iris recognition has been considered as one of the most reliable biometrics technologies in recent years [1, 2]. The human iris is the most important biometric feature candidate, which can be used for differentiating the individuals. For systems based on high quality imaging, a human iris has an extraordinary amount of unique details as illustrated in Figure.1. Features extracted from the human iris can be used to identify individuals, even among genetically identical twins [3]. Iris-based recognition system can be noninvasive to the users since the iris is an internal organ as well as externally visible, which is of great importance for the real-time applications [4].

Based on the technology developed by Daugman [3, 5, 6], iris scans have been used in several international airports for the rapid processing of passengers through the immigration which have pre registered their iris images.

A. *Proposed Method: The Main Steps*

Figure. 2 illustrates the main steps of our proposed Approach. First the image preprocessing step perform the localization of The pupil, detects the iris boundary, and isolates the collarette region, which is regarded as one of the most important areas of the iris complex pattern.

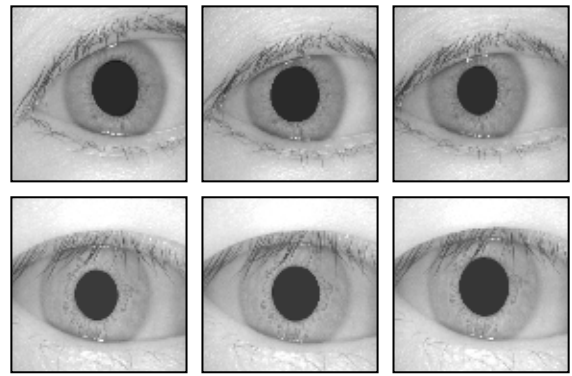

Figure. 1. Samples of iris images from CASIA [7]

The collarette region is less sensitive to the pupil dilation and usually unaffected by the eyelids and the eyelashes [8]. We also detect the eyelids and the eyelashes, which are the main sources of the possible occlusion. In order to achieve the invariance to the translation and the scale, the isolated annular collarette area is transformed to a rectangular block of fixed dimension. The discriminating features are extracted from the transformed image and the extracted features are used to train the classifiers. The optimal features subset is selected using several methods to increase the matching accuracy based on the recognition performance of the classifiers.

*B. Related works*

The usage of iris patterns for the personal identification began in the late 19th century; however, the major investigations on iris recognition were started in the last decade. In [9], the iris signals were projected into a bank of basis vectors derived by the independent component analysis, and the resulting projection coefficients were quantized as Features. A prototype was proposed in [10] to develop a 1D representation of the gray-level profiles of the iris. In [11], biometrics based on the concealment of the random kernels and the iris images to synthesize a minimum average correlation energy filter for iris authentication were formulated. In [5, 6, 12], the Multiscale Gabor filters were used to demodulate the texture phase structure information of the iris. In [13], an iris segmentation method was proposed based on the crossed chord theorem and the collarette area.

was proposed that is capable of a detailed analysis of the eye region images in terms of the position of the iris, degree of the eyelid opening, and the shape, the complexity, and the texture of the eyelids. A directional filter bank was used in [24] to decompose an iris image into eight directional sub band outputs; the normalized directional energy was extracted as features, and iris matching was performed by computing the Euclidean distance between the input and the template feature vectors. In [25], the basis of a genetic algorithm was applied to develop a technique to improve the performance of an iris recognition system. In [26], the global texture information of iris images was used for ethnic classification. The iris representation method of [10] was further developed in [27] to use the different similarity Measures for matching.

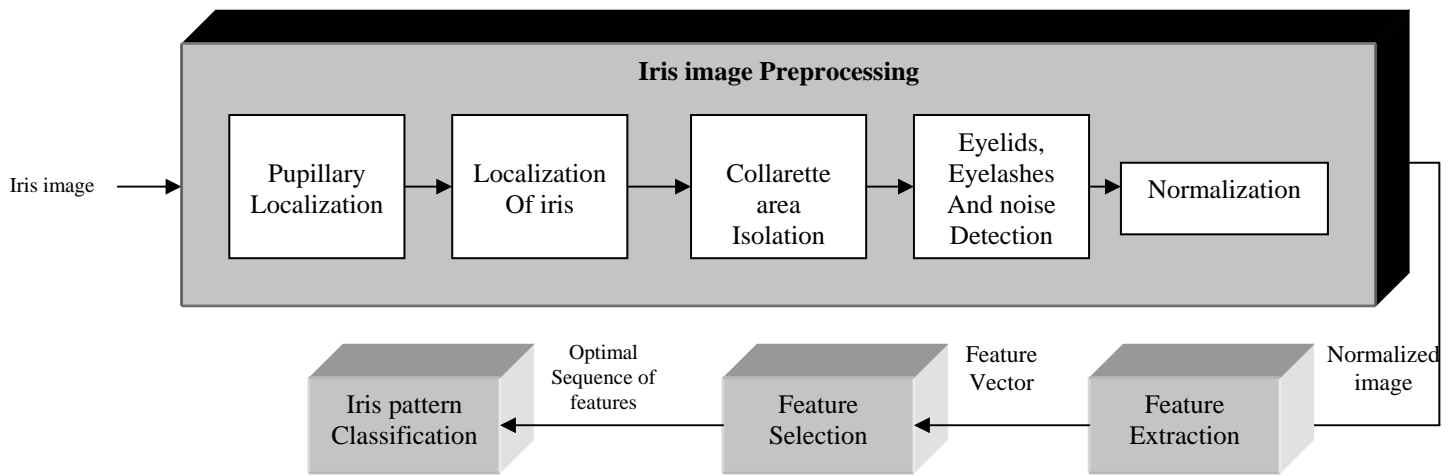

Figure.2: Flow diagram of the proposed iris recognition scheme

In [14], iris recognition technology was applied in mobile phones. In [15], correlation filters were utilized to measure the consistency of the iris images from the same eye. An interesting solution to defeat the fake iris attack based on the Purkinje image was depicted in [16]. An iris image was decomposed in [17] into four levels by using the 2D Haar wavelet transform, the fourth-level high-frequency information was quantized to form an 87-bit code, and a modified competitive learning neural network (LVQ) was adopted for classification. In [18], a modification to the Hough transform was made to improve the iris segmentation, and an eyelid detection technique was used, where each eyelid was modeled as two straight lines. A matching method was implemented in [19], and its performance was evaluated on a large dataset. In [20], a personal identification method based on the iris texture analysis was described. An algorithm was proposed for iris recognition by characterizing the key local variations in [21]. A phase-based iris recognition algorithm was proposed in [22], where the phase components were used in 2D discrete Fourier transform of iris image with a simple matching strategy. In [23], a system

Measures for matching. The iris recognition algorithm described in [28] exploited the integro differential operators to detect the inner and outer boundaries Of iris, Gabor filters to extract the unique binary vectors constituting the iris code, And a statistical matcher that analyzes the average Hamming distance between two codes. In [29], the performance of iris-based identification system was analyzed at the matching score level. A biometric system, which achieves the offline verification of certified and cryptographically secured documents called "EyeCerts" was reported in [30] for the identification of the people. An iris recognition method was used in [31] based on the 2D wavelet transform for the feature extraction and direct discriminant linear analysis for feature reduction with SVM techniques as iris pattern classifiers. In [32], an iris recognition method was proposed based on the histogram of local binary patterns to represent the iris texture and a graph matching algorithm for structural classification. An elastic iris blob matching algorithm was proposed to overcome the limitations of local feature based classifiers (LFC) in [33], and in order to recognize the various iris images properly, a novel cascading scheme was used to combine the LFC and an iris blob matcher. In [34], the authors described the determination of eye blink states by tracking the iris and the eyelids. An intensity-based iris recognition system was presented in [35], where the system exploited the local intensity changes of the visible iris textures. In [36], the iris characteristics were analyzed by using the analytic image constructed by the original image and its Hilbert transform. The binary emergent frequency functions were sampled to form a feature vector, and the Hamming distance was deployed for matching [37, 38]. In [39], the Hough transform was applied for the iris localization, a Laplacian pyramid was used to represent the distinctive spatial characteristics of the human iris, and a modified normalized correlation was applied for the matching process. In [40], various techniques have been suggested to solve the occlusion problem that happens due to the eyelids and the eyelashes. From the above discussion, we may divide the existing iris recognition approaches roughly into four major categories based on feature extraction scheme, namely, the phase-based methods [5,6,12,22], the zero-crossing representation methods [10, 27], the texture analysis-based methods [18, 21, 24, 28,39, 41–43], and the intensity variation analysis [9, 21, 44] methods. Our proposed iris recognition scheme falls in the first category. A well-established fact that the usual two-dimensional tensor product wavelet bases are not optimal for representing images consisting of different regions of smoothly varying grey-values separated by smooth boundaries. This issue is addressed by the directional transforms such as contourlets, which have the property of preserving edges. The contourlet transform is an efficient directional multiresolution image representation which differs from the wavelet transform. The contourlet transform uses non-separable filter banks developed in the discrete form; thus it is a true 2D transform, and overcomes the difficulty in exploring the geometry in digital images due to the discrete nature of the image data. The remainder of this paper is organized as follows: Section 2 deals with Iris Image Preprocessing. Section 3 deals with Feature Extraction method discussion. Section 4 deals with feature subset selection and vector creation techniques, Section 5 shows our experimental results and finally Section 6 concludes this paper.

## II. IRIS IMAGE PREPROCESSING

First, we outline our approach, and then we describe further details in the following subsections. The iris is surrounded by the various non relevant regions such as the pupil, the sclera, the eyelids, and also noise caused by the eyelashes, the eyebrows, the reflections, and the surrounding skin [9].We need to remove this noise from the iris image to improve the iris recognition accuracy.

### A. Iris / Pupil Localization

The iris is an annular portion of the eye situated between the pupil (inner boundary) and the sclera (outer boundary). Both the inner boundary and the outer boundary of a typical iris can be taken as approximate circles. However, the two circles are usually not concentric [20, 21].

### B. Eyelids, Eyelashes, and noise detection

(i) Eyelids are isolated by first fitting a line to the upper and lower eyelids using the linear Hough transform. A second horizontal line is then drawn, which intersects with the first line at the iris edge that is closest to the pupil [45].

(ii) Separable eyelashes are detected using 1D Gabor filters, since a low output value is produced by the convolution of a separable eyelash with the Gaussian smoothing function. Thus, if a resultant point is smaller than a threshold, it is noted that this point belongs to an eyelash.

(iii) Multiple eyelashes are detected using the variance of intensity, and if the values in a small window are lower than a threshold, the centre of the window is considered as a point in an eyelash as shown in Figure .3.

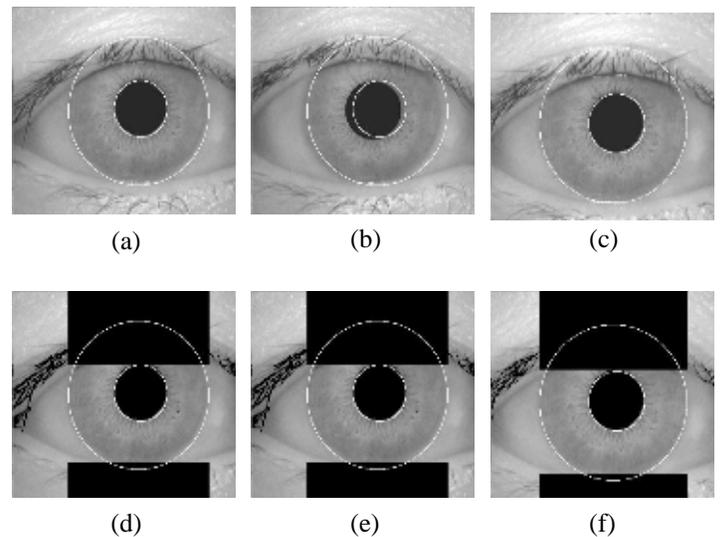

Figure.3: CASIA iris images (a), (b), and (c) with the detected Collarette area and the corresponding images (d), (e), and (f) after Detection of noise, eyelids, and eyelashes.

### C. Iris Normalization

We use the rubber sheet model [12] for the normalization of the isolated collarette area. The center value of the pupil is considered as the reference point, and the radial vectors are passed through the collarette region. We select a number of data points along each radial line that is defined as the radial resolution, and the number of radial lines going around the collarette region is considered as the angular resolution. A constant number of points are chosen along each radial line in order to take a constant number of radial data points, irrespective of how narrow or wide the radius is at a particular angle. We build the normalized pattern by backtracking to find the Cartesian coordinates of data points from the radial and angular positions in the normalized pattern [3, 5, 6]. The normalization approach produces a 2D array with horizontal dimensions of angular resolution, and vertical dimensions of radial resolution form the circular-shaped collarette area (See Figure.4I). In order to prevent non-iris region data from corrupting the normalized representation, the data points, which occur along the pupil border or the iris border, are discarded. Figure.4II (a), (b) shows the normalized images after the isolation of the collarette area.

### III. FEATURE EXTRACTION AND ENCODING

Only the significant features of the iris must be encoded so that comparisons between templates can be made. Gabor filter and wavelet are the well-known techniques in texture analysis [5, 20, 42, 46, 47]. In wavelet family, Haar wavelet [48] was applied by Jafer Ali to iris image and they extracted an 87-length binary feature vector. The major drawback of wavelets in two-dimensions is their limited ability in capturing Directional information. The contourlet transform is a new extension of the wavelet transform in two dimensions using Multi scale and directional filter banks.

sparse image expansion by applying a multi-scale transform followed by a local directional transform. It gathers the nearby basis functions at the same scale into linear structures. In essence, a wavelet-like transform is used for edge (points) detection, and then a local directional transform for contour segments detection. A double filter bank structure is used in CT in which the Laplacian pyramid (LP) [50] is used to capture the point discontinuities, and a directional filter bank (DFB) [51] to link point discontinuities into linear structures. The combination of this double filter bank is named pyramidal directional filter bank (PDFB) as shown in Figure.5.

### B. Benefits of Contourlet Transform in the Iris Feature Extraction

To capture smooth contours in images, the representation should contain basis functions with variety of shapes, in particular with different aspect ratios. A major challenge in capturing geometry and directionality in images comes from The discrete nature of the data; the input is typically sampled images defined on rectangular grids.

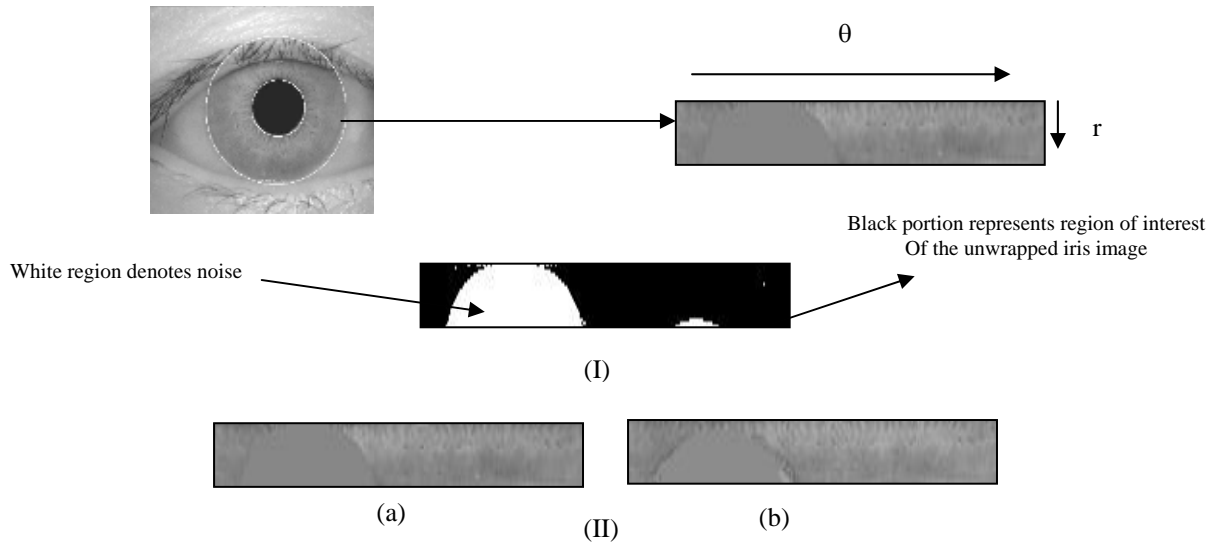

Figure.4 : (I) shows the normalization procedure on CASIA dataset; (II) (a), (b) show the normalized images of the isolated collarette regions

The feature representation should have information enough to classify various irises and be less sensitive to noises. Also in the most appropriate feature extraction we attempt to extract only significant information, more over reducing feature vector dimensions, the processing lessened and enough information is supplied to introduce iris feature vectors classification.

### A. Contourlet Transform

Contourlet transform (CT) allows for different and flexible number of directions at each scale. CT is constructed by combining two distinct decomposition stages [49], a multiscale decomposition followed by directional decomposition. The grouping of wavelet coefficients suggests that one can obtain a

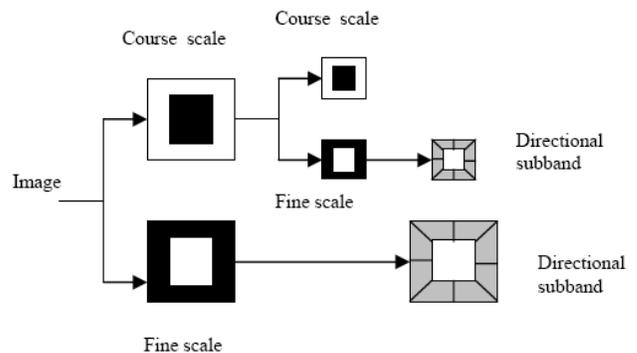

Figure. 5: Two Level Contourlet Decomposition [49]

Because of pixelization, the smooth contours on sampled images are not obvious. For these reasons, unlike other transforms that were initially developed in the continuous domain and then discretized for sampled data, the new approach starts with a discrete-domain construction and then investigate its convergence to an expansion in the continuous-domain. This construction results in a flexible multi-resolution, local, and directional image expansion using contour segments. Directionality and anisotropy are the important characteristics of contourlet transform. Directionality indicates that having basis function in many directions, only three direction in wavelet. The anisotropy property means the basis functions appear at various aspect ratios where as wavelets are separable functions and thus their aspect ratio is one. Due to this properties CT can efficiently handle 2D singularities, edges in an image. This property is utilized in this paper for extracting directional features for various pyramidal and directional filters.

### C. The Best Bit in an Iris Code

Biometric systems apply filters to iris images to extract information about iris texture. Daugman's approach maps the filter output to a binary iris code. The fractional Hamming distance between two iris codes is computed and decisions about the identity of a person are based on the computed distance. The fractional Hamming distance weights all bits in an iris code equally. However, not all the bits in an iris code are equally useful. For a given iris image, a bit in its corresponding iris code is defined as "fragile" if there is any substantial probability of it ending up a 0 for some images of the iris and a 1 for other images of the same iris. According to [52] the percentages of fragile bits in each row of the iris code, Rows in the middle of the iris code (rows 5 through 12) are the most consistent (See Figure. 6.)

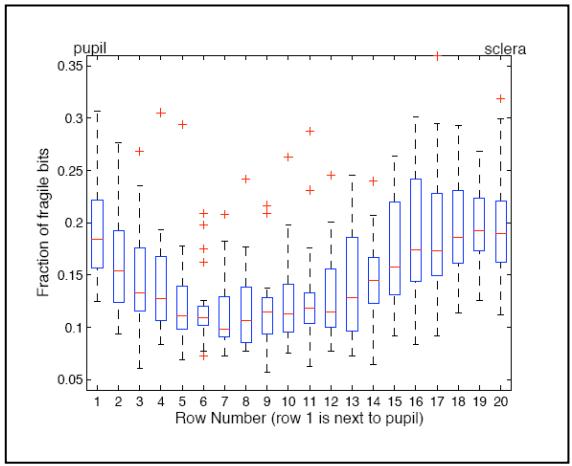

Figure.6: Percent of Fragile Bit in Iris Pattern [52]

## IV. FEATURE SUBSET SELECTION AND VECTOR CREATION IN PROPOSED METHODS

It is necessary to select the most representative feature sequence from a features set with relative high dimension [53]. In this paper, we propose several methods to select the optimal Set of features, which provide the discriminating information to classify the iris patterns. In this section we describe several methods that proposed for optimal feature selection and vector creation .also According to the method mentioned in section III*A*, we concluded the middle band of iris normalized images have more important information and less affected by fragile bits, so for introducing iris feature vector based on contourlet transform the rows between 5 and 12 in iris normalize image are decomposed into eight directional sub-band outputs using the DFB at three different scales and extract their coefficients.

### A. Gray Level Co-occurrence Matrix (GLCM)

In this method we use using the Grey Level Co-occurrence Matrix (GLCM) [54]. The technique uses the GLCM of an image and it provides a simple approach to capture the spatial relationship between two points in a texture pattern. It is calculated from the normalized iris image using pixels as primary information. The GLCM is a square matrix of size G * G, where G is the number of gray levels in the image. Each element in the GLCM is an estimate of the joint probability of a pair of pixel intensities in predetermined relative positions in the image. The $(i, j)^{th}$ element of the matrix is generated by finding the probability that if the pixel location $(x, y)$ has gray level $I_i$ then the pixel location $(x+dx, y+dy)$ has a gray level intensity $I_j$. The dx and dy are defined by considering various scales and orientations. Various textural features have been defined based on the work done by Haralick [56]. These features are derived by weighting each of the co-occurrence matrix values and then summing these weighted values to form the feature value. The specific features considered in this research are defined as follows:

1) Energy $= \sum_i \sum_j p(i,j)^2$

2) Contrast $= \sum_{n=0}^{N_g-1} n^2 \left[ \sum_{i=1}^{N_g} \sum_{j=1}^{N_g} P(i,j) \| i-j \| = n \right]$

3) Correlation $= \dfrac{\sum_i \sum_j (ij)P(i,j) - \mu_x \mu_y}{\sigma_x \sigma_y}$

4) Homogeneity $= \sum_i \sum_j \dfrac{1}{1+(i-j)^2} P(i,j)$

5) Autocorrelation $= \sum_i \sum_j (ij)P(i,j)$

6) Dissimilarity = $\sum_i \sum_j |i-j|.P(i,j)$

7) Inertia = $\sum_i \sum_j (i-j)^2 P(i,j)$

Here $\mu_x, \mu_y, \sigma_x, \sigma_y$ are mean and standard deviation along x and y axis. For creating iris feature vector we carried out the following steps:

1) Iris normalized image (Rows in the middle of the iris code (rows 5 through 12)) is decomposed up to level two.(for each image ,at level one , 2 and at level two , 4 sub band are created ) .
2) The sub bands of each level are put together, therefore at level one a matrix with 4*120 elements, and at level two a matrix with 16*120 elements is created. We named these matrixes: Matrix1 and Matrix 2.
3) By putting together Matrix1 and Matrix 2, a new matrix named Matrix3 with 20*120 elements is created. The co-occurrence of these three matrixes with offset one pixel and angles 0, 45, 90 degree is created and name this matrix: CO1, CO2 and CO3.in this case for each image 3 co-occurrence matrixes with 8*8 dimensions are created.
4) According to the Haralick's [55] theory the co-occurrence matrix has 14 properties , of which in iris biometric system we used 7 properties which are used for 3 matrixes , so the feature vector is as follow:

```
F=[ En1,Cont1,cor1,hom1,Acor1,dis1,ine1,
En2,Cont2,cor2, hom2,Acor2,dis2,ine2
En3,Cont3,cor3,hom3,Acor3,dis3,ine3]
```

In other word the feature vector in our method has only 21 elements. Also for improving results, for each sub bands and scale we create a feature vector by using GLCM.in other words for each eight sub bands in level 3 of Contourlet transform we computed GLCM properties, separately and then by combining these properties the feature vector is created. In this case the feature vector has 56 elements.

### B. Combination of Local and Global Properties in an Iris Image

Another method we used for creating iris feature vector is local and global properties of an iris image. The detailed changes of in an iris images is called the local properties. For example edges are considered a local property. The edge should be extracted from the lower levels because in upper levels the edges are usually removed. Another point is that the first level is usually very sensitive to noise. By studying coefficients, we find that when the edges are noticed, the coefficient is Positive and when the edges are gone this coefficient is Negative. After running contourlet transform the extremum value which could be positive or negative is reached. For coding these values we use the rules below:

Positive Local Maximum = 1
Negative Local Maximum = -1
Other = 0

In other words in extracted image of iris by Contourlet transform the black points represent 0, white points 1 and Grayscale points -1 .the feature vector length in this method is composed of 2520 elements.

Global properties represent the global structure of image; therefore rotation and noise do not affect them. In this method by using 12 sub band (in level 2 and 3 of Contourlet transform) we extracted the global properties and Computed average and variance for each sub bands. The feature vector in this method is as follows:

```
F=[
a1,v1,a2,v2,a3,v3,a4,v4,a5,v5,a6,v6,a7,
v7,a8,v8,a9,v9,a10,v10,a11,v11,a12,v12]
```

The feature vector in this method has only 24 elements. By combining local and global properties not only does the system with stand more noise but the system dispenses with several comparisons for overcoming the effect of rotation iris images. According to [57,58]in combination system first , the local feature vector is create, then in the matching step , a value representing the similarity amount between the input local code and local code of the class in question given in the data bank is decided. If the distance falls outside the local properties domain the global properties are extracted and distance of global code and the global code of the class in question is compared and decision is made according to the pre-decided threshold. In this technique the feature vector has 2544 elements.

### C. The Creation of Iris Feature Vector by Using PCA and ICA

In this method by using the generated sub bands and PCA (Principal Component Analysis) and ICA (Independent Component Analysis) techniques the features in question are extracted.PCA is a classic method for analyzing statistical data, extracting features and condensing data. This method via modifying data presents an appropriate representation with smaller dimensions and less added information .in this method, coordinate axes are defined in such a way that data mapping has the highest variance. To implement this method, we made use of article [59] and the features axis is developed according to level 3 sub bands of Contourlet Transform. In this method the feature vector has 1100 elements.ICA is also a statistical method for finding the components of multi variable data .Now ever, what makes this method distinct with regard to other methods of data representation is that it searches for statistically independent components that are at the same time comprised of Non-Gaussian distribution. In actual fact this method can be regarded as an extension of methods such as PCA and FA (Factor Analysis), which is by comparison stronger and in many cases of classic methods insufficient, is of much use. Similar to PCA method, we used level 3 sub bands of Contourlet Transform for creating iris feature vector. For implement ICA method we use technique hat describe in [60]. The feature vector in our method by using ICA has 1100 element similar to PCA.

### D. Feature Vector in the Coefficient domain

One of the most common methods for creating features vector is using is using the extracted coefficients by using various transformations such as Gabor filters, Wavelet etc.douagman made use of this technique in his method. In our proposed method in this section according to the extracted coefficients in level 2 of Contourlet transform, feature vector is created. Also techniques regarding decreasing vector dimensions are used.

*1) Binary vector creation with coefficient:* as stated in the previous section level 2 sub bands are extracted and according to the Following Rule are modified into binary mode:

*If Coeff (i)>=0 then NewCoeff (i) =1*
*Else NewCoeff (i) =0*

And with hamming distance between the vectors of the generated coefficients is calculated. Numbers ranging from 0 to 0.5 for inter-class distribution and 0.45 and 0.6 for intra-class distribution are included. In total 192699 comparisons inter-class and 1679 comparisons intra-class are carried out. In Figure.7 you can see inter-class and intra-class distribution. In implementing this method, we have used point 0.42 the inter-class and intra-class separation point.

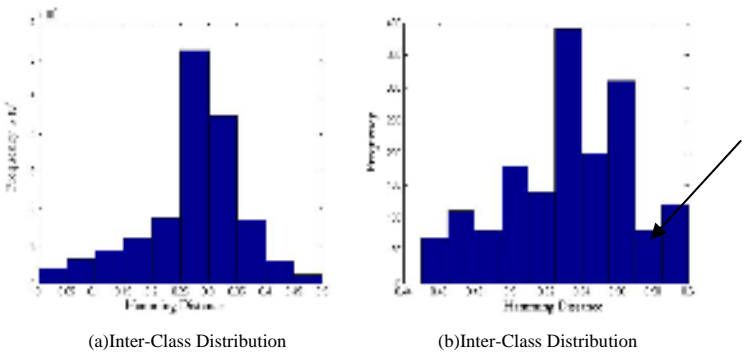

(a)Inter-Class Distribution    (b)Inter-Class Distribution

Figure.7: Inter and Intra Class Distribution

*2) Non Linear Approximation Coefficients (NLAC):* in this method we use non linear approximation coefficients for select the significant coefficient from the binary feature vector we create in the last section .for this purpose use the following formula:

$$Nsignif = round(npixel * 2.5/100) \quad (1)$$

Where npixel is the number of pixel in iris normalized image and nsignif is the number of significant coefficient. In other words it is proved that [61] only by having 2.5% of coefficients can reconstruct the image. The feature vector in our method has only 48 elements.

*3) Genetic Algorithm (GA):* optimal features subset selection with the aid of genetic algorithm is studied in this section. In fact, for creating the iris feature vector we use the level 2 binary coefficients and by using GA we try to reduce the dimensions of iris feature vector. In this method, we use MOGA [53] to select the optimal set of features, which provide the discriminating information to classify the iris patterns. In this subsection, we present the choice of a representation for encoding the candidate solutions to be manipulated by the GAs, and each individual in the population represents a candidate solution to the feature subset selection problem. If *m* be the total number of features available to choose to represent from the patterns to be classified (*m* = 600 in our case), the individual is represented by a binary vector of dimension, *m*. If a bit is a 1, it means that the corresponding feature is selected, otherwise the feature is not selected (See Figure.8) this is the simplest and most straightforward representation scheme [53]. In this work, we use the roulette wheel selection [53], which is one of the most common and easy to implement selection mechanism. Usually, there is a fitness value associated with each chromosome, for Example, in a minimization problem, a lower fitness value means that the chromosome or solution is more optimized to the problem while, a higher value of fitness indicates a less optimized chromosome. Our problem consists of optimizing two objectives:

(i) Minimization of the number of features,
(ii) Minimization of the recognition error rate of the Classifier.

Therefore, we deal with the multi objectives optimization Problem. In Table I you can see the parameters used in the genetic algorithm.

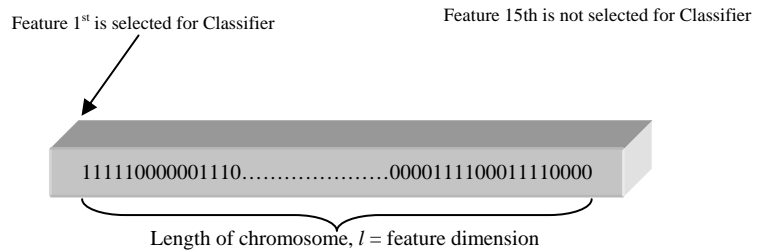

Figure.8: Binary feature vector of *l* dimension.

Table. I: GA Parameters

| Parameters | CASIA Dataset |
|---|---|
| Population size | 108 (the scale of iris sample) |
| Length of chromosome code | 600 (selected dimensionality of feature sequence) |
| Crossover probability | 0.65 |
| Mutation probability | 0.002 |
| Number of generation | 110 |

E. *Average Absolute Deviation (ADD):* In this algorithm, the feature value is the average absolute deviation (AAD) of each output image defined as follows:

$$F = \frac{1}{N}\left[\sum_{N}|f(x,y)-m|\right] \quad (2)$$

where *N is* the number of pixels in the image, *m* is the mean of the image, and *f(x,y)* is the value at point (*x, y*). The AAD feature is a statistic value similar to variance, but experimental results show that the former gives slightly better performance than the latter. The average absolute deviation of each filtered image constitutes the components of our feature vector. These Features are arranged to form a 1D feature vector of length 1280 for each input image. (160 elements for each sub bands in lev 3 of Contourlet transform).

## V. EXPRIMENTAL RESULT

To evaluate the performance of this proposed system we use "CASIA"[7] iris image database (version 1) created by National Laboratory of pattern recognition, Institute of Automation, Chinese Academy of Science that consists of 108 subjects with 7 sample each. Images of "CASIA" iris image database are mainly from Asians. For each iris class, images are captured in two different sessions. The interval between two sessions is one month. There is no overlap between the training and test samples. In our experiments, three-level Contourlet decomposition is adopted. The above experiments are performed in Matlab 7.0. The normalized iris image obtained from the localized iris image is segmented by Dugman method. We have used the filters designed by A. Cohen, I. Daubechies, and J.-C. Feauveau. For the quincunx filter banks in the DFB stage. In Table II we compared our proposed methods with some other well known methods from 3 view points: feature vector length, the correct of percentage classification and feature extraction time. Also we modified the classifier of well known method to SVM for better comparison.

Table II: Comparison Pertaining to Our methods and
And some well - known Method.

| Method | The Feature Vector Length(Bit) | Classifier | | The Correct Of Percentage Classifier (%) | | Feature Extraction(ms) |
|---|---|---|---|---|---|---|
| *Well Known methods* | | | | | | |
| Daugman[3] | 2048 | HD | SVM | 100 | 100 | 628.5 |
| Lim[17] | 87 | LVQ | SVM | 90.4 | 92.3 | 180 |
| Jafar Ali[48] | 87 | HD | SVM | 92.1 | 92.8 | 260.3 |
| Ma[20] | 1600 | ED | SVM | 95.0 | 95.9 | 80.3 |
| *Our Proposed Methods* | | | | | | |
| Gray Level Co-occurrence Matrix | | | | | | |
| GLCM[54] | 21 | SVM | | 94.2 | | 20.3 |
| GLCM (Combining Sub bands) | 56 | SVM | | 96.3 | | 20.3 |
| Local and Global Feature | | | | | | |
| Local Feature | 2520 | HD | | 90.32 | | 20.3 |
| Global Feature | 24 | ED | | 78.6 | | 20.3 |
| Combining Local and Global Feature | 2544 | HD+ED | | 93.2 | | 20.3 |
| Binary Vector in coefficient domain | | | | | | |
| Binary Vector | 2520 | HD | | 96.5 | | 20.3 |
| NLAC[55] | 48 | SVM | | 91.3 | | 20.3 |
| GA | 600 | SVM | | 97.81 | | 20.3 |
| Other methods | | | | | | |
| ADD | 1280 | SVM | | 92.63 | | 20.3 |
| PCA | 1100 | SVM | | 90 | | 20.3 |
| ICA | 1100 | SVM | | 85.9 | | 20.3 |

### A. Discussion

- Using GLCM causes the feature vector with appropriate dimensions and acceptable classification accuracy.
- The highest acceptable classification accuracy percentage is arrived at in GA.
- The feature vector in the coefficients domain, PCA, ICA and ADD method is easily implemented.
- Using the global and local properties do not come to good results in isolation while a combination of these two is really noise resistant and leads us to good results.
- NLAC has really appropriate dimensions for feature vector and acceptable classification accuracy percentage.
- All the proposed methods in this paper save time in the processing and extraction of feature in comparison with the known existence methods.

## VI. CONCLUSION

In this paper we proposed an effective algorithm for iris feature extraction using contourlet transform. For reduce iris feature vector we use several techniques. For Segmentation and normalization we use Daugman methods. The Contourlet transform is used to extract the discriminating features, and several methods are applied for the feature subset selection. Our proposed methods can classify iris feature vector properly. The rate of expected classification for the fairly large number of experimental date in this paper verifies this claim. In other words most proposed methods in this paper provide a less feature vector length with an insignificant reduction of the percentage of correct classification.

AUTHORS PROFILE

**Amir Azizi:** Graduated From Islamic Azad University, Mashhad Branch and Received the B.S. in Computer Engineering .He received the M.S. degree in Artificial intelligence From the Islamic Azad University, Qazvin Branch**.**

**Hamid Reza Pourreza:** Graduated From Ferdowsi University of Mashhad in EE and received the PHD in C.E From AmirKabir University of Technology. He is Assistant Professor of Computer Eng. Dept. School of Engineering*,* Ferdowsi University of Mashhad.